\documentclass[10pt,twocolumn,letterpaper]{article}

\usepackage{cvpr}              
\definecolor{cvprblue}{rgb}{0.21,0.49,0.74}
\usepackage[pagebackref,breaklinks,colorlinks,allcolors=cvprblue]{hyperref}
\usepackage{hyperref}
\usepackage{url}
\usepackage{times}
\usepackage{epsfig}
\usepackage{graphicx}
\usepackage{amsmath}
\usepackage{amssymb}
\usepackage{makecell}
\usepackage{multirow}
\usepackage{bbding}
\usepackage{xcolor}
\usepackage{tabularx}
\usepackage{subcaption}
\usepackage{booktabs} 
\usepackage[linesnumbered,ruled,vlined]{algorithm2e}

\def\confName{CVPR}
\def\confYear{2026}

\title{UNIV: Unified Foundation Model for Infrared and Visible Modalities}

\author{
\textbf{Fangyuan Mao}$^{1,2}$\thanks{Equal Contribution}\hspace{9pt}
\textbf{Shuo Wang}$^{1,2}$\footnotemark[1]\hspace{9pt}
\textbf{Jilin Mei}$^{1,2}$\thanks{Corresponding Author}\hspace{9pt} 
\textbf{Shun Lu}$^{1,2}$\hspace{9pt}\\
\textbf{Chen Min}$^{1,2}$\hspace{9pt}
\textbf{Fuyang Liu}$^{1,2}$\hspace{9pt}
\textbf{Xiaokun Feng}$^{3}$\hspace{9pt}
\textbf{Meiqi Wu}$^{3}$\hspace{9pt}
\textbf{Yu Hu}$^{1,2}$\footnotemark[2] \\
$^{1}$Research Center for Intelligent Computing Systems, CAS ICT\\
$^{2}$University of Chinese Academy of Sciences\\
$^{3}$Institute of Automation, Chinese Academy of Sciences\\
}

\begin{document}
\maketitle
\begin{abstract}

Joint RGB–infrared perception is essential for achieving robustness under diverse weather and illumination conditions. Although foundation models excel within single modalities, they suffer from substantial cross-modal degradation—an issue we attribute to a pattern shortcut, i.e., a modal bias that prioritizes superficial sensor patterns over underlying semantics.
To address this problem, we introduce \textbf{UNIV}, a \textbf{UN}ified foundation model for \textbf{I}nfrared and \textbf{V}isible modalities. At the core of UNIV lies \textbf{Patch Cross-modal Contrastive Learning (PCCL)}, a self-supervised contrastive learning strategy that constructs a unified cross-modal feature space. PCCL employs a frozen pretrained model to sample pseudo patch pairs based on semantic similarity, and aligns infrared–visible representations by attracting semantically related pairs while repelling unrelated ones. This process simultaneously enhances cross-modal alignment and inter-class semantic separability, guiding the model to focus on semantic structure rather than falling into pattern shortcuts.
To further enable cross-modal learning, we introduce MVIP, the most comprehensive visible–infrared benchmark to date, containing 98,992 precisely aligned image pairs across diverse scenes. Extensive experiments demonstrate UNIV’s superior performance on infrared tasks (+1.7 mIoU for semantic segmentation and +0.7 mAP for detection), while maintaining competitive accuracy on RGB tasks. Our code is available at \href{https://github.com/fangyuanmao/UNIV}{https://github.com/fangyuanmao/UNIV}.

\end{abstract}    
\section{Introduction}
\label{sec:introduction}

\begin{figure}[ht]
    \centering
    \includegraphics[width=1\linewidth]{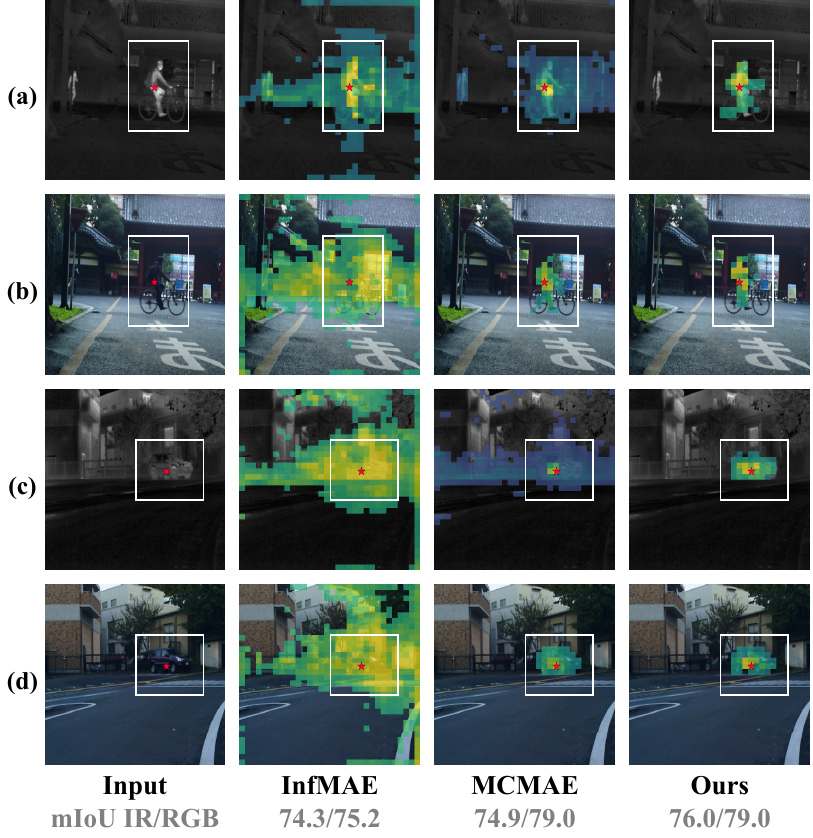}
    \caption{\textbf{Visualization of attention maps from the final layer of different pre-trained models for a specific patch (marked with a star).} (a) and (c) show a cyclist and a vehicle in the infrared modality, while (b) and (d) show the same objects in the visible RGB modality. Our UNIV establishes excellent cross-modal alignment and inter-class separability, enabling consistent attention localization for focused regions (marked with white boxes) at the patch level. \textcolor{gray}{\textbf{mIoU}} denotes mIoU scores for infrared (IR) and visible (RGB) semantic segmentation on the MSRS~\cite{Tang2022PIAFusion}.}
    \label{fig:intro}
    \vspace{-15pt}
\end{figure}



Recent advances in pre-trained foundation models have achieved impressive results in both visible~\cite{caron2021emerging, chen2021empirical, he2022masked, gao2022convmae} and infrared~\cite{liu2025infmae, zhang2023pad, zhang2025unip} modalities. However, these models are trained in isolation, and a significant cross-domain gap remains when applying them across modalities. This gap limits their generalization under diverse perception settings, including RGB-only, IR-only, and RGB–IR joint scenarios. A unified encoder that produces semantically consistent features for both modalities is therefore essential.

Existing single-modality pre-training suffers from a \textbf{pattern shortcut}: models rely heavily on modality-specific cues—such as color and brightness in RGB or thermal intensity in IR—instead of semantic structure. \cref{fig:intro}(b) shows that IR-trained models like InfMAE~\cite{liu2025infmae} transfer brightness-sensitive attention to RGB images, causing incorrect localization. Conversely, visible-trained encoders such as MCMAE~\cite{gao2022convmae} fail in IR scenes (\cref{fig:intro}(c)) due to the absence of color and texture. These observations reveal a fundamental limitation: \textbf{cross-modal alignment and semantic separability cannot be achieved by single-modality pre-training alone}.

\begin{figure}[ht]
    \centering
    \includegraphics[width=1\linewidth]{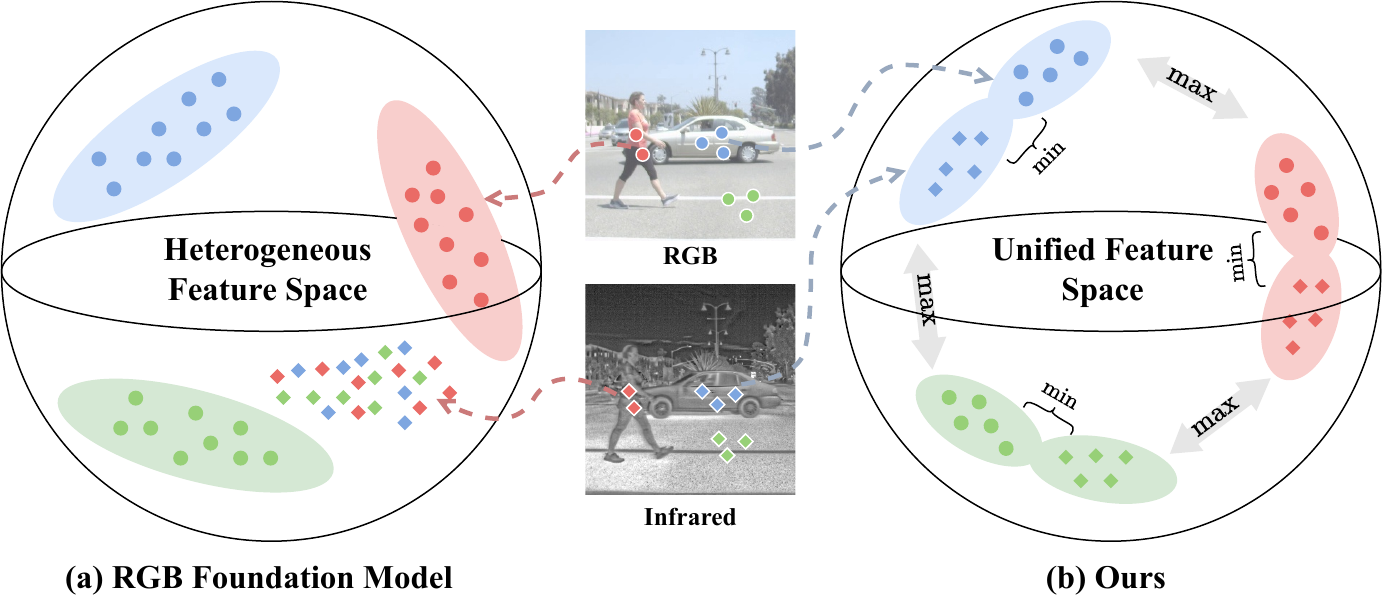}
    \caption{\textbf{Proposed Unified Feature Space.} \textbf{UNIV} learns a semantically rich and robust feature mapping by promoting intra-class cross-modal alignment and enhancing inter-class separation.}
    \label{fig:keyidea}
    \vspace{-15pt}
\end{figure}

As shown in \cref{fig:keyidea}, achieving unified RGB–IR representations requires shaping a feature space where same-class samples cluster together and different-class samples are well separated. Supervised learning can impose this structure but depends heavily on large labeled RGB–IR datasets. Importantly, the geometry we need—attraction between semantically related samples and repulsion between unrelated ones—is inherently the geometry that contrastive learning builds. Contrastive learning constructs such spaces by optimizing positive and negative relations directly, without needing ground-truth labels. Leveraging this property, we extract pseudo positive and negative patch pairs from a pre-trained encoder and use contrastive learning to refine the shared representation space for RGB–IR alignment.

In this paper, we introduce \textbf{UNIV}, a \textbf{UN}ified foundation model for \textbf{I}nfrared and \textbf{V}isible modalities. UNIV uses \textbf{Patch-level Cross-modal Contrastive Learning (PCCL)} to close the feature gap while strengthening inter-class semantics. We leverage the semantic priors encoded in a frozen ImageNet-pretrained RGB model: its last-layer attention map provides patch-level similarity, which we binarize into pseudo positive and negative labels. Features from this frozen encoder serve as anchors. During training, either an RGB image or its paired IR image is fed into our learnable encoder, and contrastive learning aligns its representations with the anchors according to the pseudo labels. This process pulls semantically relevant patches together across modalities and pushes apart irrelevant ones, producing a unified and discriminative feature space.

To support this cross-modal learning, we introduce the Multi-scene Visible Infrared Pairs Dataset (MVIP), a comprehensive collection of 98,992 precisely aligned visible–infrared image pairs curated from five public datasets~\cite{hwang2015multispectral,Flirdataset,li2021lasher,sun2022drone,jia2021llvip}. Covering urban driving, surveillance, and aerial drone scenarios, MVIP provides a solid foundation for robust cross-modal research (see Appendix A for details).

Our key contributions are:
\begin{itemize}
    \item \textbf{Patch Cross-modal Contrastive Learning}: A self-supervised method that adaptively reduces the cross-modal feature gap while enhancing inter-class discriminability.
    \item \textbf{MVIP Dataset}: The largest visible–infrared benchmark with 98,992 aligned image pairs spanning diverse scenes.
    \item \textbf{State-of-the-Art Performance}: Our model achieves top results on infrared tasks, including semantic segmentation (+1.7 mIoU on MSRS) and object detection (+0.7 mAP on M3FD), while preserving the downstream performance of RGB tasks.
\end{itemize}

\section{Related Works}
\label{sec:relatedwork}

\subsection{The Foundation Model}

The success of pre-trained models like BERT~\cite{devlin2018bert}, ViTs~\cite{dosovitskiy2020image}, and CLIP~\cite{radford2021learning} demonstrate their ability to learn general representations from vast datasets. Following this paradigm, infrared-specific models (e.g., PAD~\cite{zhang2023pad}, InfMAE~\cite{liu2025infmae}) have also been developed. Yet, a critical gap persists: existing efforts are characterized by modal specialization, employing distinct architectures and pre-training objectives for RGB vs. infrared. This has resulted in a lack of a unified foundation model that generalizes across modality, severely limiting practical application.


\subsection{Contrastive Learning}
Self-supervised representation learning has progressed rapidly in visual recognition. Contrastive learning~\cite{chopra2005learning} remains the core paradigm. DINO~\cite{caron2021emerging} leverages teacher–student distillation for label-free global features, and DINOv2~\cite{oquab2023dinov2} scales this framework for highly transferable representations. iBOT~\cite{zhou2021ibot} introduces patch-level tokenization to improve dense prediction. Recent methods push toward finer granularity and larger-scale pretraining: I-JEPA~\cite{assran2023self} predicts masked patch features, SigLIP~\cite{zhai2023sigmoid} adopts a sigmoid objective for image–text training, and ViC-MAE~\cite{hernandez2024vic} integrates masked autoencoding with contrastive learning across images and videos.


\subsection{Cross Modal Alignment}
Cross-modal alignment has evolved from static image–text matching to video, temporal sequences, and spectral modalities. CLIP~\cite{radford2021learning} established a shared latent space through contrastive pretraining, while subsequent work introduced fine-grained and hierarchical video–text alignment across frames, clips, and phrases~\cite{wray2019fine, nian2022multi, ye2023hitea, jiang2022tencent}. Recent methods further explore modality-specific anchors and multi-granularity interactions for more robust alignment\cite{ma2022x, chen2023tagging}. These advances motivate our visible–infrared cross-modal feature alignment.

\begin{figure*}[t]
    \centering
    \includegraphics[width=1\linewidth]{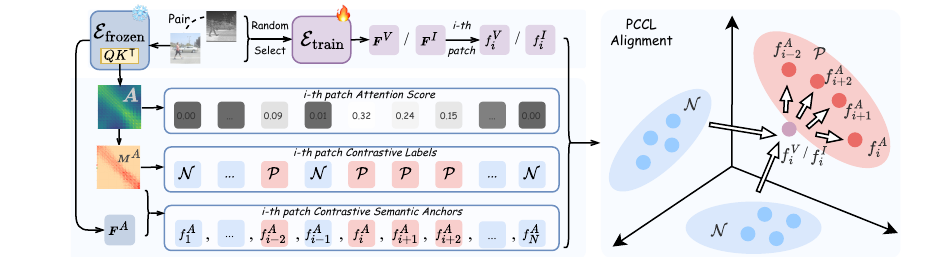}
    \caption{\textbf{Overview of the proposed UNIV.} Attention maps from a frozen visible RGB backbone are transformed to similarity pseudo-labels $\boldsymbol{M}^{A}$. The cross-modality similarity map $\boldsymbol{M}^{IA}$ and visible modality similarity map $\boldsymbol{M}^{VA}$ are optimized to align with $\boldsymbol{M}^{A}$.}
    \label{fig:overview}
\end{figure*}

\section{Methodology}
\label{sec:methodology}

\subsection{Core Idea}

As shown in \cref{fig:keyidea}, our core idea is to learn a new feature mapping $\mathcal{E}$ that projects input images into a more discriminative feature space, where features of the same semantic class remain close across modalities, while features of different classes are pushed apart. We formulate the problem as follows:

Given paired visible RGB and infrared images, we extract their features using a trainable encoder $\mathcal{E}_\text{train}$ on the top of Fig.\ref{fig:overview}:
\begin{align}
    \boldsymbol{F}^{V} &= \mathcal{E}_{\text{train}}(\text{Img}_{\text{Visible}}), \\
    \boldsymbol{F}^{I} &= \mathcal{E}_{\text{train}}(\text{Img}_{\text{Infrared}}) \in \mathbb{R}^{d_k \times N},
\end{align}
where $d_k$ denotes the feature dimension and $N$ represents the number of image patches. The infrared feature $\boldsymbol{F}^{I}$ can be decomposed as $\boldsymbol{F}^{I} = \{f_1^{I}, f_2^{I}, \dots, f_N^{I}\}$, where each $f_i^I$ corresponds to the feature of the $i$-th patch. Similarly, $\boldsymbol{F}^{V}$ can be decomposed in the same manner.

To guide the feature learning, we propose the \textbf{Attention-Guided Patch Anchors} method, which extracts robust positive and negative labels from attention maps and generates patch contrastive semantic anchors $\boldsymbol{F}^{A}$ defined in Sec.\ref{sec:AGPA}. Based on these anchors, we design an effective \textbf{Patch Cross-modal Contrastive Learning (PCCL)} optimization strategy to refine the feature space.

\subsection{Attention-Guided Patch Anchors}
\label{sec:AGPA}

In preliminary experiments, we observed that features extracted by a frozen pre-trained RGB encoder $\mathcal{E}_{\text{frozen}}$ already exhibit approximate clustering behavior. This inspires us to use this pre-existing feature space as a reference, where patch features serve as semantic anchors. This avoids re-optimizing the entire feature space and accelerates training.

Specifically, we leverage the last-layer attention map from $\mathcal{E}_{\text{frozen}}$ applied to the visible modality. As illustrated in \cref{fig:intro} (b) and (d), these attention maps show high correlations for semantically similar regions and low correlations for dissimilar ones. We obtain the last-layer attention map $\boldsymbol{A}$ from $\mathcal{E}_\text{frozen}$:
\begin{equation}
\label{eq:atten_map}
\boldsymbol{A} = \text{softmax}\left(\frac{QK^T}{\sqrt{d_k}}\right) \in \mathbb{R}^{N \times N},
\end{equation}
where $Q$ and $K$ are query and key matrices, and $d_k$ is the feature dimension.

We then convert $\boldsymbol{A}$ into a binary pseudo-label matrix $\boldsymbol{M}^A$ using the following three rules:
\begin{itemize}
    \item Self-similarity enforcement: Set all diagonal elements $\boldsymbol{M}^A_{ii} = 1$ to preserve patch identity.
    \item Correlation selection: For each row $i$, sort the elements $\{\boldsymbol{M}^A_{ik}\}_{k=1}^N$ in descending order. Let $\pi(\cdot)$ be the mapping from sorted indices to original indices such that $\boldsymbol{M}^A_{i\pi(1)} \geq \cdots \geq \boldsymbol{M}^A_{i\pi(N)}$. Find the smallest $m$ satisfying $\sum_{k=1}^m \boldsymbol{M}^A_{i\pi(k)} > \gamma$, and set $\boldsymbol{M}^A_{i\pi(1)}, \dots, \boldsymbol{M}^A_{i\pi(m)} = 1$.
    \item Dissimilarity suppression: Set all non-selected entries to 0.
\end{itemize}

This process is formally expressed as:
\begin{equation}
\label{eq:Pij}
\boldsymbol{M}^{A}_{ij} = \begin{cases} 
1 & \text{if } i = j, \\
1 & \text{if } j \in \left\{\pi(1), \dots, \pi(m_\text{thr})\right\}, \\
0 & \text{otherwise},
\end{cases}
\end{equation}
where $m_\text{thr} = \min\left\{ m \, \middle| \, \sum_{k=1}^m \boldsymbol{M}^{A}_{i\pi(k)} > \gamma \right\}$.

Thus, we could partition the frozen robust RGB features into patch contrastive semantic anchors $\boldsymbol{F}^{A} = \{f_1^{A}, f_2^{A}, \dots, f_N^{A}\}$ in term of pseudo-label $\boldsymbol{M}^A$.




    
    
    

\subsection{Patch Cross-modal Contrastive Learning}

To reconstruct the unified feature space associated with the PCCL alignment shown on the right side of Fig.\ref{fig:overview}, We first discuss the optimization for the inter-class separation of infrared modality. Given the semantic anchors $\boldsymbol{F}^{A}$, we define the following objective:

Let $\mathcal{P}$ denote positive sample pairs (i.e., patches belonging to the same class), and $\mathcal{N}$ denote negative sample pairs (i.e., patches from different classes). Let $\mathcal{S}$ be a dissimilarity function, and $\tau$ be a temperature hyper-parameter. A common choice for $\mathcal{S}$ is cosine similarity:
\begin{equation}
    \mathcal{S}(z_i, z_j) = \frac{z_i^\top z_j}{\|z_i\|\|z_j\|}.
\end{equation}
The optimization goal is:
\begin{equation}
\min \left( \mathbb{E}_{(i,j)\sim \mathcal{N}} \left[ \mathcal{S}(f_i^{A}, f_j^{I}) / \tau \right] - \mathbb{E}_{(i,j)\sim \mathcal{P}} \left[ \mathcal{S}(f_i^{A}, f_j^{I}) / \tau \right] \right).
\label{eq:op1}
\end{equation}
Define the similarity matrix:
\begin{equation}
\label{eq: Fiv}
M^{IA}_{ij} = \mathcal{S}(f^A_{i}, f^I_{j}) / \tau, \quad \boldsymbol{M}^{IA} = \{M^{IA}_{ij}\}_{i,j=1}^N.
\end{equation}
Using the pseudo-label matrix $\boldsymbol{M}^A$, we can rewrite \cref{eq:op1} as:
\begin{align}
&\min\ \mathbb{E}_{M^{A}_{ij}=0}[M_{ij}^{IA}] - \mathbb{E}_{M^{A}_{ij}=1}[M_{ij}^{IA}] \\
\Leftrightarrow &\min\ \mathcal{L}_{\text{disc}} = \sum_{M^{A}_{ij}=0} M_{ij}^{IA} - \sum_{M^{A}_{ij}=1} M_{ij}^{IA}.
\label{eq:op2}
\end{align}

We propose a loss function $\mathcal{L}_{IA}$ that minimizes the KL divergence between $\boldsymbol{M}^A$ and $\sigma(\boldsymbol{M}^{IA})$, where $\sigma(\cdot) = \frac{1}{1 + e^{-(\cdot)}}$ is the sigmoid function. When a patch has multiple positive samples, optimizing the KL divergence is equivalent to optimizing a binary cross-entropy (BCE) loss between $\boldsymbol{M}^A$ and $\sigma(\boldsymbol{M}^{IA})$ as shown below:
\begin{equation}
    \begin{aligned}
\mathcal{L}_{IA} &\equiv D_{\text{KL}}(\boldsymbol{M}^A \| \sigma(\boldsymbol{M}^{IA})) \\
&= \sum_{i=1}^{N} \sum_{j=1}^{N} D_{\text{KL}}(M^A_{ij} \| \sigma(M^{IA}_{ij}))\\
&= \sum_{i=1}^{N} \sum_{j=1}^{N} \mathcal{L}_{\text{BCE}}(M^A_{ij}, \sigma(M^{IA}_{ij}))+H(\boldsymbol{M}^A)\\
&\propto \mathcal{L}_{\text{BCE}}(\boldsymbol{M}^A, \sigma(\boldsymbol{M}^{IA})),
\label{eq:op3}
\end{aligned}
\end{equation}
where $H(\boldsymbol{M}^A)$ is the entropy of $\boldsymbol{M}^A$, which is constant.

Thus, we could see that the KL divergence $\mathcal{L}_{IA}$ is equal to $\mathcal{L}_{\text{BCE}}(\boldsymbol{M}^A, \sigma(\boldsymbol{M}^{IA}))$ defined in \cref{eq:bce_def} plus the entropy term $H(\boldsymbol{M}^A)$, which is a constant with respect to the model parameters. Therefore, minimizing the KL divergence is equivalent to minimizing the BCE loss.
\begin{equation}
    \begin{aligned}
\mathcal{L}_{\text{BCE}}(\boldsymbol{M}^A, \sigma(\boldsymbol{M}^{IA})) &= -\sum_{i=1}^{N}\sum_{j=1}^{N}  \Big( M^A_{ij} \log\big(\sigma(M^{IA}_{ij})\big) \\
&+ (1 - M^A_{ij}) \log\big(1 - \sigma(M^{IA}_{ij})\big) \Big).
\label{eq:bce_def}
\end{aligned}
\end{equation}


Despite BCE is defined on the sigmoid output $\sigma(M^{IA}_{ij})$, its optimization direction is fully consistent with the discriminative objective in \cref{eq:op2}. For positive pairs ($M^{A}_{ij}=1$), minimizing BCE increases $\sigma(M^{IA}_{ij})$, which requires increasing the similarity $M^{IA}_{ij}$. For negative pairs ($M^{A}_{ij}=0$), minimizing BCE reduces $\sigma(M^{IA}_{ij})$, pushing $M^{IA}_{ij}$ toward smaller values. This behavior directly matches \cref{eq:op2}, which enlarges positive similarities while suppressing negative ones. Consequently, BCE provides a stable optimization surrogate for the same contrastive direction defined by $\mathcal{L}_{\text{disc}}$. Through this equivalence, we optimize the infrared features to adaptively align with the semantic anchors in the feature space.

To further enhance the cross-modal alignment, we extend this approach to the RGB modality. Since we use paired images, the semantics of RGB and infrared modalities should be highly aligned. For RGB input, we similarly define $\mathcal{L}_{VA} \equiv D_{\text{KL}}(\boldsymbol{M}^A \| \sigma(\boldsymbol{M}^{VA}))$ to refine RGB features using the same semantic anchors. In practice, we randomly select the input modality (RGB or IR). Since \textbf{both modalities are optimized toward the same semantic anchors, they gradually align in the feature space}, achieving cross-modal alignment. Finally, the total optimization loss function employed in PCCL training is defined as follows
The total loss combines both objectives:  
\begin{equation}  
\mathcal{L}_\text{PCCL} = \alpha \cdot \mathcal{L}_{IA} + \beta \cdot \mathcal{L}_{VA},  
\label{eq:lpccl}
\end{equation}  
where $\alpha$ and $\beta$ are the weight of $\mathcal{L}_{IA}$ and $\mathcal{L}_{VA}$ .

\section{Experiments}
\label{sec:experiments}
\subsection{Experimental Setup}  

\begin{table*}[ht]
\setlength\tabcolsep{2pt}
\centering
\caption{\textbf{The performance of UNIV on infrared semantic segmentation and object detection tasks.} MSRS-IR and M3FD-IR refer to the infrared sub-datasets of MSRS~\cite{Tang2022PIAFusion} and M3FD~\cite{liu2022target}, respectively. \textbf{Total} params denotes the encoder's total parameter count, while \textbf{Pre-train} params represents the number of parameters updated during pre-training. Our model achieves state-of-the-art performance compared to existing infrared pre-trained models. The top two results are highlighted in \textbf{bold} and \underline{underlined}.}
\label{tab:ir-downstream}
\begin{tabular}{lcccc|lcccc}
\toprule
\multirow{2}{*}{\textbf{\begin{tabular}[c]{@{}c@{}}Semantic\\ Segmentation\end{tabular}}} & \multirow{2}{*}{\textbf{\begin{tabular}[c]{@{}c@{}}Pre-train\\ Dataset\end{tabular}}} & \multirow{2}{*}{\textbf{\begin{tabular}[c]{@{}c@{}}Total/Pre-train\\ Params(M)\end{tabular}}} & \multicolumn{2}{c|}{\textbf{MSRS-IR}} & \multirow{2}{*}{\makecell[c]{\textbf{Object}\\\textbf{Detection}}} & \multirow{2}{*}{\textbf{\begin{tabular}[c]{@{}c@{}}Pre-train\\ Dataset\end{tabular}}} & \multirow{2}{*}{\textbf{\begin{tabular}[c]{@{}c@{}}Total/Pre-train\\ Params(M)\end{tabular}}} & \multicolumn{2}{c}{\textbf{M3FD-IR}} \\ \cmidrule{4-5} \cmidrule{9-10} 
 &  &  & \textbf{mIoU} & \textbf{mAcc} &  &  &  & \textbf{mAP} & \textbf{AP50} \\ \midrule 
\multicolumn{10}{l}{\textcolor{gray}{Training from scratch}} \\
\textbf{UperNet}~\cite{xiao2018unified} & None & 25.6/- & 65.6 & 74.7 & \textbf{YoloV8}~\cite{yolov8} & None & 46.0/- & 40.8 & 63.0 \\ 
\textbf{DeeplabV3+}~\cite{chen2018encoder} & None & 25.6/- & 65.2 & 73.8 & \textbf{FastRCNN}~\cite{ren2016faster} & None & 44.6/- & 25.2 & 45.6 \\ 
\textbf{FPN}~\cite{lin2017feature} & None & 25.6/- & 62.9 & 71.4 & \textbf{DINO}~\cite{caron2021emerging} & None & 85.0/- & 45.3 & 75.3 \\
\textbf{ISANet}~\cite{huang2019interlaced} & None & 25.6/- & 67.8 & 74.9 & \textbf{DETR}~\cite{carion2020end} & None & 60.0/- & 46.7 & 79.0 \\
\textbf{InfMAE}~\cite{liu2025infmae} & None & 88.7/- & 61.1 & 61.1 & \textbf{InfMAE}~\cite{liu2025infmae} & None & 88.7/- & 48.7 & 80.0 \\ \midrule
\multicolumn{10}{l}{\textcolor{gray}{Pre-training on infrared dataset}} \\
\textbf{Vanila MAE}~\cite{he2022masked} & Inf30 & 85.8/85.8 & 71.4 & 78.2 & \textbf{Vanila MAE}~\cite{he2022masked} & Inf30 & 85.8/85.8 & 51.4 & 83.4 \\ 
\textbf{MCMAE}~\cite{gao2022convmae} & Inf30 & 88.7/88.7 & 72.1 & 79.8 & \textbf{MCMAE}~\cite{gao2022convmae} & Inf30 & 88.7/88.7 & 55.7 & 88.4 \\ 
\textbf{InfMAE}~\cite{liu2025infmae} & Inf30 & 88.7/88.7 & 74.3 & 82.5 & \textbf{InfMAE}~\cite{liu2025infmae}& Inf30 & 88.7/88.7 & 56.2 & 88.1 \\ \midrule
\multicolumn{10}{l}{\textcolor{gray}{Pre-training on visible RGB dataset}} \\
\textbf{MCMAE}~\cite{gao2022convmae} & IN1K & 88.7/88.7 & 74.9 & 83.8 & \textbf{MCMAE}~\cite{gao2022convmae} & IN1K & 88.7/88.7 & \underline{56.6} & \underline{88.5} \\ \midrule
\multicolumn{10}{l}{\textcolor{gray}{Pre-training on cross-modality dataset}} \\
\textbf{UNIV(Full)} & \scriptsize{IN1K+MVIP} & 88.7/88.7 & \textbf{76.6} & \textbf{85.1} & \textbf{UNIV(Full)} & \scriptsize{IN1K+MVIP} & 88.7/88.7 & 56.2 & 88.4 \\ 
\textbf{UNIV(LoRA)} & \scriptsize{IN1K+MVIP} &88.7/1.8 & \underline{76.0} & \underline{84.6} & \textbf{UNIV(LoRA)} & \scriptsize{IN1K+MVIP} & 88.7/1.8 & \textbf{56.9} & \textbf{88.7} \\ \bottomrule
\end{tabular}
\end{table*}

\subsubsection{Datasets}
To facilitate PCCL training, we construct the \textbf{Multi-scene Visible-Infrared Pair (MVIP)} dataset, comprising 197,984 aligned visible-infrared image pairs (98,992 per modality) curated from five publicly available benchmarks~\cite{hwang2015multispectral,Flirdataset,li2021lasher,sun2022drone,jia2021llvip}. MVIP spans diverse real-world scenarios, including urban driving, surveillance, and aerial drone views, featuring common objects such as vehicles and pedestrians, making it suitable for cross-modal perception in various environments. To reduce redundancy in sequential data, we apply frame downsampling to ensure a representative subset of samples. Detailed dataset construction procedures are provided in the Datasets section of the Appendix.

\subsubsection{Implementation Details}
\noindent\textbf{Pre-training:}
We initialize models (both $\mathcal{E}_\text{train}$ and $\mathcal{E}_\text{frozen}$) with an ImageNet-1K (IN1K)~\cite{deng2009imagenet} pre-trained MCMAE~\cite{gao2022convmae} (ViT-B backbone), leveraging its robust feature and attention map, and train $\mathcal{E}_\text{train}$ branch using PCCL on the MVIP dataset to enhance unified feature extraction. We also use LoRA to finetune the backbone. LoRA adapters with a default rank of 8 and a dropout rate of 0.1 are integrated into the ViT’s feed-forward layers, QKV linear transformations, and projection layers. We train 400 epochs for the LoRA integration version and 200 epochs for full parameter pre-training with a batch size of 256. The default loss coefficients are set to $\alpha = \beta = 1$, with $\gamma = 0.6$ and temperature $\tau = 0.04$. Further training details can be found in Implementation Details section of Appendix.

\subsection{Comparison with State-of-the-art Infrared Pre-trained Models}
To evaluate the effectiveness of our proposed UNIV, we conduct comprehensive comparisons with several baseline methods. These include models trained from scratch, models pre-trained solely on large-scale infrared datasets, and models pre-trained solely on visible RGB dataset.

We evaluate our model on two infrared sub-datasets: MSRS~\cite{Tang2022PIAFusion} for semantic segmentation and M3FD~\cite{liu2022target} for object detection, using the UperNet~\cite{xiao2018unified} head for semantic segmentation and MaskRCNN~\cite{he2017mask} head for object detection.

The results of \textcolor{gray}{Training from scratch} and \textcolor{gray}{Pre-training on infrared dataset} are from \cite{liu2025infmae}. All methods are trained to convergence under identical settings to ensure a fair comparison. Details and visualization can be found in Appendix.

\begin{table*}[ht]
\centering
\setlength\tabcolsep{2pt}
\caption{\textbf{Comparison of mIoU with different pre-trained backbones using UperNet.} Pre-training datasets are annotated with their respective modalities. InfMAE fails to converge on ADE20K dataset, showing as N/A.}
\label{tab:RGB-downstream}
\begin{tabular}{lcccc}
\toprule
    \textbf{Methods} & \textbf{Pre-train Dataset}  & \textbf{ADE20K(RGB)} & \textbf{MSRS(IR)} & \textbf{MSRS(RGB)} \\ \midrule
    \textbf{MAE}~\cite{he2022masked} & IN1K(RGB) & 48.1 & 71.4 & 74.2 \\ 
    \textbf{InfMAE}~\cite{liu2025infmae} & Inf30(IR) & N/A & 74.3 & 75.2 \\ 
    \textbf{MCMAE}~\cite{gao2022convmae} & IN1K(RGB) & \textbf{51.7} & \underline{74.9} & \textbf{79.0} \\ \midrule
    \textbf{UNIV(LoRA)} & IN1K+MVIP(RGB+IR) & \underline{51.2} (\textcolor{gray}{-0.5}) & \textbf{76.0}(\textcolor{blue}{+1.1}) & \textbf{79.0} \\
     \bottomrule
\end{tabular}
\end{table*}

\subsubsection{Semantic Segmentation}
The results of infrared image semantic segmentation on the MSRS-IR dataset are shown in \cref{tab:ir-downstream}. A key observation is that models trained from scratch consistently underperform compared to pre-trained models, underscoring the importance of pre-training for learning modality-invariant semantic features. For example, MCMAE pre-trained on IN1K achieves a 2.8 mIoU gain over the same model pre-trained on Inf30. This performance gap highlights the critical role of strong pre-training in building robust infrared representations.

Building upon this observation, our proposed UNIV model achieves state-of-the-art performance on MSRS-IR. With LoRA-based pre-training, UNIV reaches 76.0 mIoU, surpassing the previous SOTA InfMAE by 1.7 mIoU while updating only 2\% of parameters. The full-parameter variant further improves to 76.6 mIoU. These results validate the effectiveness of our PCCL and LoRA pre-training strategy in balancing accuracy and efficiency, and they demonstrate that, within the unified feature space, the inter-class semantic separability of infrared features is significantly enhanced, leading to substantial downstream improvements.

\subsubsection{Object Detection}
The results of infrared object detection on the M3FD-IR dataset are presented in \cref{tab:ir-downstream}. When fine-tuned with LoRA, the proposed UNIV model achieves 56.9 mAP and 88.7 AP50, surpassing the previous state-of-the-art InfMAE by 0.7 mAP and 0.6 AP50, demonstrating its strong detection capability. Interestingly, cross-modality fine-tuning of MCMAE pre-trained on IN1K also produces competitive performance, suggesting that single-modality pre-training can still capture partially shared cross-modal features.

In contrast, the full-parameter pre-training variant of UNIV yields suboptimal results (56.2 mAP, lower than MCMAE-IN1K’s 56.6 mAP). Our analysis indicates that full-parameter UNIV tends to over-optimize patch similarity during pre-training, at the expense of preserving semantic richness within individual patches. This limitation is particularly pronounced in object detection, where overlapping bounding boxes lead to highly entangled patch-level information. LoRA fine-tuning, by applying low-rank updates, better maintains inter-patch semantics while selectively adjusting patch relationships, enabling more effective learning for detection tasks.

\subsection{UNIV on RGB Downstream Task}
To evaluate the inter-class seperation ability on the RGB modal, we compare our UNIV model performance with several baseline methods. As shown in \cref{tab:RGB-downstream}, we evaluate the downstream semantic segmentation performance on the ADE20K~\cite{zhou2017scene} and MSRS-RGB~\cite{Tang2022PIAFusion} datasets. All methods are trained under identical settings to ensure a fair comparison. Details can be found in the Implementation Details section of Appendix.

\begin{table*}[ht]
\setlength\tabcolsep{2pt}
\centering
\caption{\textbf{Comparison of mIoU with different training paradigms on SODA-IR~\cite{li2020segmenting} and MFNet-IR~\cite{ha2017mfnet} datasets.} \textbf{Total} params denotes the encoder's total parameter count, while \textbf{Training} params represents the number of parameters updated during additional pre-training on MSIP~\cite{zhang2023pad}.}
\label{tab:finetune}
\begin{tabular}{lccccccc}
\toprule
    \multirow{2}{*}{\textbf{Methods}} & \multirow{2}{*}{\makecell[c]{\textbf{Pre-train}\\\textbf{Dataset}}} & \multirow{2}{*}{\makecell[c]{\textbf{Pre-train}\\\textbf{Method}}} & \multirow{2}{*}{\textbf{Epoch}} & \multirow{2}{*}{\makecell[c]{\textbf{Total / Training}\\\textbf{Params(M)}}} & \textbf{SODA-IR} & \multicolumn{2}{c}{\textbf{MFNet-IR}} \\
    \cmidrule{7-8}
     & & & & & test & val & test \\ \midrule
    \multicolumn{2}{l}{\textcolor{gray}{Full pre-training from scratch}} &  &  &  &  &  & \\
    \textbf{MAE~\cite{he2022masked}} & MSIP & MIM & 400 & 85.8 / 85.8 & 61.46 & 39.58 & 42.98 \\ 
    \midrule
    \multicolumn{2}{l}{\textcolor{gray}{Cross-domain fine-tuning}} &  &  &  &  &  & \\ 
    \textbf{MoCo v3~\cite{chen2021empirical}} & IN1K & CL & 300 & 85.8 / - & 67.05 & 43.23 & 47.60 \\ 
    \textbf{MILAN~\cite{hou2022milan}} & IN1K & MIM & 400 & 85.8 / - & 68.93 & 45.08 & 47.11 \\ 
    \textbf{MAE~\cite{he2022masked}} & IN1K & MIM & 1600 & 85.8 / - & 67.51 & 44.47 & 47.35 \\ 
    \textbf{MCMAE~\cite{gao2022convmae}} & IN1K & MIM & 1600 & 88.7 / - & \underline{69.57} & \underline{50.29} & \underline{50.31} \\ 
    \midrule
    \multicolumn{2}{l}{\textcolor{gray}{Full pre-training from IN1K}} &  &  &  &  &  & \\
    \textbf{MAE~\cite{he2022masked}} & IN1K+MSIP & MIM & 1600 & 85.8 / 85.8 & 64.90 & 43.97 & 45.36 \\ 
    \midrule
    \multicolumn{2}{l}{\textcolor{gray}{Pre-training with extra params}} &  &  &  &  &  & \\
    \textbf{PAD~\cite{zhang2023pad}} & IN1K+MSIP & MIM & 100 & 87.0 / \textbf{1.2} & 68.41 & 46.89 & 48.82 \\ 
    \textbf{UNIV(LoRA)} & IN1K+MSIP & PCCL & 100 & 88.7 / \underline{1.8} & \textbf{69.60} & \textbf{50.78} & \textbf{51.06} \\

    \bottomrule
\end{tabular}
\end{table*}

The heterogeneous feature space from Infrared-specific InfMAE encounters modality collapse in visible RGB downstream tasks, highlighting the challenges of cross-modal adaptation. However, UNIV, producing an unified semantic feature space, maintains near-baseline RGB performance while enabling efficient infrared adaptation. On the MSRS-RGB dataset, UNIV maintains original competitive performance (79.0 mIoU), matching the baseline and further validating its effectiveness in cross-modal tasks. On ADE20K, UNIV achieves 51.2 mIoU — retaining 99\% of the original MCMAE (IN1K-pretrained) capability. These results collectively demonstrate that UNIV effectively constructs a unified semantic feature space, enabling robust cross-modal transfer while preserving strong performance on the original modality.

\subsection{Comparison with Fine-tuning based Methods}

To further evaluate the effectiveness of our fine-tuning strategy, we compare UNIV with PAD~\cite{zhang2023pad}, an adapter-based fine-tuning method built on MIM pre-training. For a fair comparison, UNIV is pre-trained on a subset of PAD’s MSIP dataset (only 46\% infrared images) and fine-tuned on SODA-IR~\cite{li2020segmenting} and MFNet-IR~\cite{ha2017mfnet} under the same settings.

As shown in \cref{tab:finetune}, PAD with an MAE backbone confirms that cross-domain pre-training is more advantageous than direct cross-domain fine-tuning or training from scratch. However, UNIV trained with our PCCL paradigm achieves higher mIoU than both the MCMAE baseline~\cite{gao2022convmae} and PAD, indicating that PCCL extracts more discriminative infrared semantic features.

Crucially, the advantage of PCCL comes from constructing a unified feature space shared by visible and infrared modalities. Unlike MIM-based approaches~\cite{liu2025infmae}, which focus solely on domain-specific infrared cues—often limited to brightness and lacking color-texture semantics—PCCL explicitly aligns IR features with the richer semantic structure of visible features. This alignment retains the intrinsic inter-class geometry of the visible modality and transfers it to infrared representations. As a result, in the reconstructed unified space, UNIV preserves strong inter-class separability for infrared features, enabling more effective downstream fine-tuning even with limited IR data.

\subsection{Feature Space Analysis}

We quantitatively and qualitatively analyze our unified feature space on the MSRS dataset. We employ Earth Mover's Distance (EMD)~\cite{rubner2000earth} to assess cross-modal alignment and inter-class separation, and use t-SNE to visualize the feature distributions.

\subsubsection{Cross-modal Alignment}

\begin{figure}[htbp]
    \centering
    \includegraphics[width=1\linewidth]{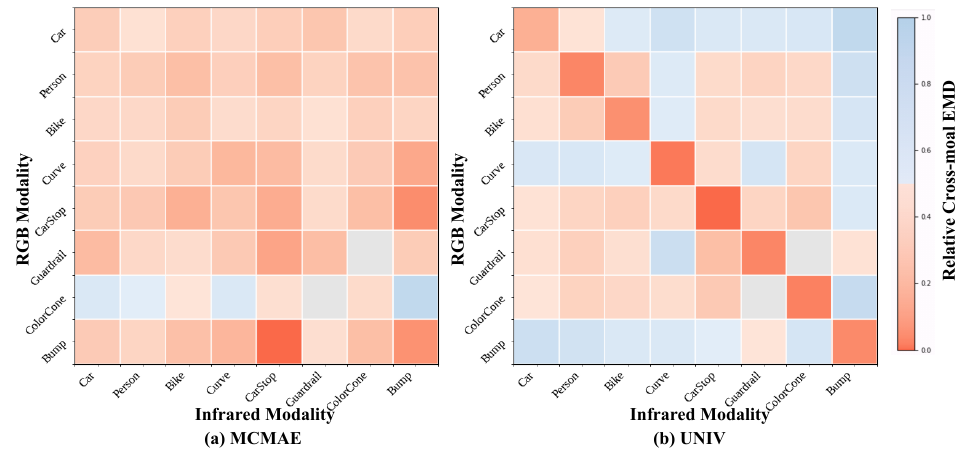}
    \caption{\textbf{Cross-modal Aligment.} Our model achieves excellent cross-modal alignment with efficient inter-class separability.}
    \label{fig:cross_modal}
    \vspace{-15pt}
\end{figure}

As shown in \cref{fig:cross_modal}, we compute the EMD distances between cross-modal semantic feature distributions. In \cref{fig:cross_modal}(a), the category-wise distance matrix between RGB and IR modalities in the unaligned MCMAE feature space shows that same-category features (diagonal entries) are not closer than those of different categories, indicating poor alignment. In contrast, \cref{fig:cross_modal}(b) demonstrates strong same-category alignment in our unified feature space, where diagonal entries exhibit significantly higher similarity than off-diagonal ones.

\subsubsection{Inter-class Separability}

\begin{figure}[htbp]
    \centering
    \includegraphics[width=1\linewidth]{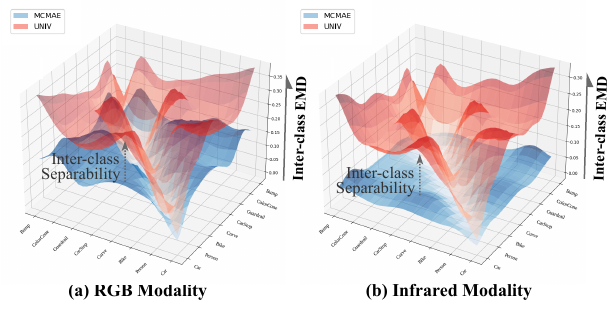}
    \caption{\textbf{Inter-class Separability.} Our model exhibits better inter-class separability.}
    \label{fig:inter_class}
\end{figure}

The inter-class separability is quantified by computing category-wise distances within each modality. A comparative analysis in \cref{fig:inter_class} reveals that our UNIV space (red surface) consistently achieves greater inter-category distances than MCMAE (blue) across all modalities. This clear improvement attests to the superior discriminative power of our contrastive learning optimization.

\subsubsection{Feature Space Visualization}

\begin{figure}[htbp]
    \centering
    \includegraphics[width=1\linewidth]{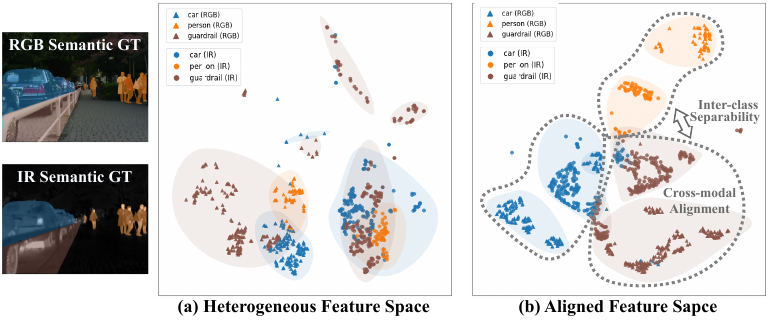}
    \caption{\textbf{Heterogeneous vs Aligned Feature Space.} Left: MCMAE projection feature space. Right: our proposed unified feature space.}
    \label{fig:tsne_vis}
    \vspace{-15pt}
\end{figure}

The t-SNE analysis in \cref{fig:tsne_vis} evaluates the realization of the unified feature spaces from \cref{fig:keyidea}. Our method successfully separates features of different semantics while clustering cross-modal counterparts. This stands in clear contrast to \cref{fig:keyidea}(a), where RGB features are partially separable but infrared features are semantically entangled, highlighting the more coherent semantic organization achieved by our approach.

\subsection{Ablation Studies}

\subsubsection{Does Unified Feature Benefits Semantic Learning?}

Catastrophic forgetting—where learning new concepts interferes with previously acquired knowledge~\cite{french1999catastrophic}—is a major challenge in cross-modal pre-training. By contrast, our reorganized unified feature space enables visible and infrared representations to be jointly structured rather than sequentially overwritten, effectively preserving modality-specific knowledge while fostering shared semantics. This unified space serves as a stable geometric scaffold where both modalities maintain their intrinsic semantic topology, thereby improving the model’s ability to learn category-discriminative features.

To verify whether such a unified space indeed benefits semantic learning, we fine-tune UNIV with a UperNet segmentation head on the MSRS-IR and MSRS-RGB datasets. As shown in \cref{tab:fulltrain}, the results provide direct evidence that unified feature learning leads to stronger semantic representation across both modalities.

\begin{table}[h]

\setlength\tabcolsep{5pt}
    \caption{\textbf{Ablation study on the proposed losses and LoRA integration.}}
    \label{tab:fulltrain}
    \centering
    \begin{tabular}{ccccccc}
    \toprule
    \textbf{} & \multirow{2}{*}{$\mathcal{L}^{IA}$} & \multirow{2}{*}{$\mathcal{L}^{VA}$} & \multirow{2}{*}{\textbf{LoRA}} & \multirow{2}{*}{\makecell[c]{\textbf{Training}\\ \textbf{Params(M)}}} & \textbf{IR} & \textbf{RGB} \\ 
    \textbf{} & ~ & ~ & ~ & ~ & \textbf{mIoU} & \textbf{mIoU} \\ \midrule
    \textbf{(a)} & ~ & ~ & ~ & - & 74.9 & \textbf{79.0} \\ 
    \textbf{(b)} & \CheckmarkBold & ~ & ~ & 88.7 & 76.0 & 77.9 \\
    \textbf{(c)} & \CheckmarkBold & \CheckmarkBold & ~ & 88.7 & \textbf{76.6} & \textbf{79.0} \\
    \textbf{(d)} & \CheckmarkBold  & ~ & \CheckmarkBold & 1.8 & 75.8 & 78.9 \\ 
    \textbf{(e)} & \CheckmarkBold & \CheckmarkBold & \CheckmarkBold & 1.8 & \underline{76.0} & \textbf{79.0} \\ 
    \bottomrule
    \end{tabular}
\end{table}

Introducing $\mathcal{L}^{IA}$ (comparing (a) and (b)) improves the infrared modality’s performance by 1.1 mIoU but slightly degrades RGB performance. This degradation suggests that aligning infrared features with visible ones may introduce minor shifts in the visible feature space. However, adding $\mathcal{L}^{VA}$ in (c) restores RGB performance to its baseline level (79.0 mIoU) while further enhancing infrared performance to 76.6 mIoU. This indicates that $\mathcal{L}^{VA}$ effectively mitigates catastrophic forgetting through natural cross-modal alignment during optimizing.

Integrating LoRA in (d) reduces the number of trainable parameters to 1.8M while maintaining competitive performance (IR: 75.8 mIoU, RGB: 78.9 mIoU). The minimal RGB drop suggests that LoRA constrains parameter updates, thereby preserving modality-specific representations. When $\mathcal{L}^{VA}$ and LoRA are combined in (e), the model achieves a balanced trade-off, reaching 76.0 mIoU for infrared and maintaining 79.0 mIoU for RGB. This result underscores the complementary nature of $\mathcal{L}^{VA}$ and LoRA with the benefit of unified feature space.

\subsubsection{Loss Functions}
To validate the effectiveness of the proposed $\mathcal{L}_\text{PCCL}$, we compare it with two commonly used loss functions: $\mathcal{L}_{\text{MSE}}$ Eq.(\ref{eq:lmse}) and $\mathcal{L}_{\text{NCE}}$ Eq.(\ref{eq:lnce}).
\begin{equation}
    \mathcal{L}_{\text{MSE} } =\| \boldsymbol{F}^{I}-\boldsymbol{F}^{A}\|_{2}^2 +\| \boldsymbol{F}^{V}-\boldsymbol{F}^{A}\|_{2}^2,
    \label{eq:lmse}
\end{equation}
\begin{equation}
    \mathcal{L}_{\text{NCE} } = \text{CE(} \boldsymbol{M}^{IA},[1,...,N]) + \text{CE(} \boldsymbol{M}^{VA},[1,...,N]),
    \label{eq:lnce}
\end{equation}
where CE denotes the Cross Entropy loss. The results, as shown in \cref{tab:losses}, highlight the advantages of PCCL over these baselines. 

\begin{table}[t]
    \caption{\textbf{Ablation study on different cross-modality losses.}}
\label{tab:losses}
    \centering
    \begin{tabular}{llcc}
    \toprule
        \multirow{2}{*}{\textbf{Loss function}} & \multirow{2}{*}{\textbf{Definition}} & \multicolumn{2}{c}{\textbf{MSRS-IR}} \\
        \cmidrule{3-4}
        ~ & & \textbf{mIoU} & \textbf{mAcc} \\ \midrule
        $\mathcal{L}_{\text{MSE}}$ & \cref{eq:lmse} & 75.6 & 84.1 \\ 
        $\mathcal{L}_{\text{NCE}}$ & \cref{eq:lnce} & 75.7 & 84.2 \\ 
        $\mathcal{L}_{\text{PCCL}}$(ours) & \cref{eq:lpccl} & \textbf{76.0} & \textbf{84.6} \\ 
        \bottomrule
    \end{tabular}

\end{table}

$\mathcal{L}_{\text{MSE}}$ minimizes the $L_2$ distance between cross-modality features, enabling rapid convergence but failing to fully exploit infrared modality-specific representations, limiting its effectiveness. $\mathcal{L}_{\text{NCE}}$ enhances feature similarity within paired instances while reducing it across non-paired ones, as shown in models like CLIP~\cite{radford2021learning}. In our implementation, $\mathcal{L}_{\text{NCE}}$ optimizes patch similarity at identical spatial locations (e.g., the main diagonal of $\boldsymbol{M}^{IA}$). However, it overlooks semantic relationships between patches at different locations, which are critical in non-iconic scenes with overlapping semantics. 

To address these limitations, we propose $\mathcal{L}_\text{PCCL}$, which considers both semantic alignment of corresponding patches and correlations among multiple similar patches, enabling more robust and discriminative feature learning. As demonstrated in \cref{tab:losses}, PCCL effectively balances modality-specific and cross-modal feature learning, offering a superior solution for cross-modal pre-training.

More ablation studies can be found in Appendix.
\section{Conclusion}
\label{sec:conclusion} 

In this work, we propose UNIV, a unified foundation model that bridges infrared and visible modalities. By integrating PCCL, UNIV reconstructs an effective unified semantic feature space that achieves both robust cross-modal alignment and strong inter-class semantic separability. Extensive experiments demonstrate UNIV’s superiority. Ablation studies further validate the efficiency of our framework and the effectiveness of the proposed PCCL-optimized unified feature space construction. Overall, UNIV offers a general and scalable solution for building a unified cross-modal semantic feature space.

\clearpage
\appendix
\setcounter{page}{1}
\maketitlesupplementary

We provide detailed descriptions of the dataset used in the paper (\cref{sec:app1}), the equivalence derivation of the optimization method (\cref{sec:app_opt}), experimental details (\cref{sec:app2}), visualization results (\cref{sec:app3}), and ablation studies (\cref{sec:app4}).

\section{Datasets}
\label{sec:app1}

\subsection{Pre-training Datasets}
Our pre-training dataset MVIP consists of 98,992 visible-infrared image pairs from the following datasets:

\noindent\noindent\textbf{FLIR-Align~\cite{FLIR_Align_github}}.The FLIR-Align dataset is a subset of the FLIR~\cite{Flirdataset} dataset, specifically filtered to ensure strict alignment between RGB and infrared image pairs. The dataset primarily features highway driving scenes. The dataset consists of 4,890 RGB-infrared pairs for training, 126 pairs for validation, and 126 pairs for testing. During pretraining, we primarily use the training set for model training.

\noindent\textbf{KAIST}~\cite{hwang2015multispectral}. The KAIST dataset, captured from a driving viewpoint, is a multimodal RGBT collection. It contains exactly 95,324 aligned image pairs, each consisting of synchronized infrared and RGB modalities. This dataset spans diverse environments such as urban streets, campus, and downtown, with data collected across both daytime and nighttime. For training purposes, we sample every fourth image, resulting in a subset of 12,538 paired images.

\noindent\textbf{DroneVehicle}~\cite{sun2022drone}. Collected from aerial drone platforms, the DroneVehicle dataset provides 28,439 precisely aligned thermal-RGB image pairs for vehicle detection. This aerial dataset covers multiple scenarios including metropolitan streets, residential zones, and parking lots, with comprehensive daytime and nighttime coverage. We employ the complete training set of 17,990 image pairs for our pre-training dataset.

\noindent\textbf{LLVIP}~\cite{jia2021llvip}. Designed for pedestrian detection in surveillance contexts, the LLVIP dataset offers 15,488 perfectly aligned thermal-visible image pairs. From this collection, we extract the full set of 12,025 training pairs for our pre-training dataset.

\noindent\textbf{LasHeR}~\cite{li2021lasher}. As a comprehensive RGBT tracking benchmark, the LasHeR dataset comprises 1,224 sequence pairs with more than 730,000 frame-level thermal-visible image pairs across various scenarios and object classes. For training, we sample every tenth frame from the 979 available training sequences, generating a set of 51,843 image pairs.

\subsection{Downstream Datasets}

\noindent\textbf{MSRS}. The MSRS~\cite{Tang2022PIAFusion} dataset consists of 1,444 pixel-level annotated image pairs, capturing urban scenes in both infrared and visible spectra. The dataset is divided into 1,083 training pairs and 361 testing pairs, with annotations for eight driving-related classes. 

\noindent\textbf{M3FD-IR}. The M3FD~\cite{liu2022target} dataset, designed for infrared and visible target detection, contains 4,200 images labeled with six target categories related to driving. We use the dataset split provided by InfMAE~\cite{liu2025infmae} where 3360 images for training and 840 images for testing.

\noindent\textbf{SODA-IR}.The SODA-IR~\cite{li2020segmenting} dataset includes a diverse range of indoor and outdoor scenes for infrared semantic segmentation. It comprises 1,168 training images and 1,000 test images, covering 20 semantic categories, including road, building, car, chair, lamp, person , and more.

\noindent\textbf{MFNet-IR}.The MFNet-IR~\cite{ha2017mfnet} dataset adopts the infrared modality of MFNet~\cite{ha2017mfnet},which is an RGB-infrared semantic segmentation dataset covering perspectives from autonomous driving vehicles. The dataset consists 1,569 training images (with 820 captured during the daytime and 749 at night), 392 validation images, and 393 test images.

\noindent\textbf{ADE20K}. ADE20K~\cite{zhou2017scene} is a widely-used semantic segmentation dataset which contains 25,562 images of 150 fine-grained categories. The dataset is split into training, validation, and testing sets. We fine-tune the different backbone with segmentation head on ADE20K training set (20210 images) and test on validation split (2000 images).

\section{Optimization}
\label{sec:app_opt}

We demonstrate that the optimization direction of our proposed cross-modal loss function aligns with the intended objective. As shown in \cref{eq:op-s1}, \cref{eq:op-s2} and \cref{eq:op-s3}, optimizing $\mathcal{L}_{IA}$ is equivalent to optimizing $\mathcal{L}_{\text{BCE}}(\boldsymbol{M}^A, \sigma(\boldsymbol{M}^{IA}))$, which in turn corresponds to optimizing a smooth approximation of $\mathcal{L}_{disc}$. By applying the sigmoid smoothing function $\sigma$ to $\boldsymbol{M}^{IA}$, we not only ensure differentiability but also promote more stable and robust training dynamics. This smoothing strategy effectively mitigates potential gradient instability and facilitates smoother convergence during optimization.

\begin{figure*}[!ht]
\begin{equation}
\label{eq:op-s1}
\begin{aligned}
\mathcal{L}_{IA} &\equiv D_{\text{KL}}(\boldsymbol{M}^A \| \sigma(\boldsymbol{M}^{IA})) \\
&= \sum_{i=1}^{N} \sum_{j=1}^{N} D_{\text{KL}}(M^A_{ij} \| \sigma(M^{IA}_{ij}))\\
&= \sum_{i=1}^{N} \sum_{j=1}^{N} \left[ M^A_{ij} \log \frac{M^A_{ij}}{\sigma(M^{IA}_{ij})} + (1 - M^A_{ij}) \log \frac{1 - M^A_{ij}}{1 - \sigma(M^{IA}_{ij})} \right] \\
&= \sum_{i=1}^{N} \sum_{j=1}^{N} \underbrace{\left[-M^A_{ij}\log\left(\sigma(M^{IA}_{ij})\right)-(1-M^A_{ij})\log\left(1-\sigma(M^{IA}_{ij})\right)\right]}_{\mathcal{L}_{\text{BCE}}\left(M^A_{ij}, \sigma(M^{IA}_{ij})\right)}+\underbrace{\left[M^A_{ij}\log(M^A_{ij})+(1-M^A_{ij})\log(1-M^A_{ij})\right]}_{H({M}^A_{ij})}\\
&= \sum_{i=1}^{N} \sum_{j=1}^{N} \mathcal{L}_{\text{BCE}}\left(M^A_{ij}, \sigma(M^{IA}_{ij})\right)+\underbrace{H(\boldsymbol{M}^A)}_{\text{const}}\\
&\propto \mathcal{L}_{\text{BCE}}(\boldsymbol{M}^A, \sigma(\boldsymbol{M}^{IA})).
\end{aligned}
\end{equation}
\end{figure*}

\begin{figure*}[!ht]
    \begin{equation}
    \label{eq:op-s2}
    \begin{aligned}
    \because \min \mathcal{L}_{\text{BCE}}\left(\boldsymbol{M}^A, \sigma(\boldsymbol{M}^{IA})\right) 
    &\Leftrightarrow \min \sum_{i=1}^{N} \sum_{j=1}^{N} -M^A_{ij}\log\left(\sigma(M^{IA}_{ij})\right)-(1-M^A_{ij})\log\left(1-\sigma(M^{IA}_{ij})\right)\\
    &\Leftrightarrow \min \sum_{M^{A}_{ij} = 0} -\log\left(1-\sigma(M^{IA}_{ij})\right) - \sum_{M^{A}_{ij} = 1} \log\left(\sigma(M^{IA}_{ij})\right)\\
    &\Leftrightarrow \min \sum_{M^{A}_{ij} = 0, \sigma(M^{IA}_{ij})\rightarrow 0 } \sigma(M^{IA}_{ij}) - \sum_{M^{A}_{ij} = 1, \sigma(M^{IA}_{ij})\rightarrow 1}\sigma(M^{IA}_{ij})\\
    &\Leftrightarrow \min \sum_{M^{A}_{ij} = 0} \sigma(M^{IA}_{ij}) - \sum_{M^{A}_{ij} = 1}\sigma(M^{IA}_{ij}).
    \end{aligned}
\end{equation}
\end{figure*}

\begin{figure*}[!ht]
    \begin{equation}
    \label{eq:op-s3}
    \therefore \min \mathcal{L}_{IA} \Leftrightarrow \min \mathcal{L}_{\text{BCE}}\left(\boldsymbol{M}^A, \sigma(\boldsymbol{M}^{IA})\right)
    \Leftrightarrow \min \mathcal{L}_{disc} = \sum_{M^{A}_{ij} = 0} M_{ij}^{IA} - \sum_{M^{A}_{ij} = 1} M_{ij}^{IA}.
\end{equation}
\end{figure*}

\section{Implementation Details}
\label{sec:app2}

\subsection{Backbone}
In default, we leverage the visible RGB pre-trained model MCMAE~\cite{gao2022convmae} as our backbone model. MCMAE has hierarchical feature output, we utilize the final feature map with PCCL training.

\subsection{Training Schedule}

\subsubsection{Pre-training}
All pre-training experiments are conducted using PyTorch on 8 NVIDIA V100 GPUs. The weights of all models pre-trained on ImageNet are sourced from MCMAE~\cite{gao2022convmae}. The default configuration is detailed in \cref{tab:pretraining}. \textit{To keep the same pre-train dataset as PAD~\cite{zhang2023pad}, for MFNet and SODA experiments, we only apply the DroneVehicle, LLVIP and LasHeR dataset (which are the subset of MSIP used in PAD) in the pre-training.}

\subsubsection{LoRA Configuration}
We employ the LoRA technique with the assistance of PEFT~\cite{peft}. The configuration for LoRA training is provided in \cref{tab:lora-config}.

\subsubsection{Fine-tuning}
\noindent\textbf{Object Detection.} Following InfMAE~\cite{liu2025infmae}, we use Detectron2~\cite{wu2019detectron2} to train the detection model. All experiments are conducted on 4 NVIDIA V100 GPUs. The default configuration is provided in \cref{tab:obd}.

\noindent\textbf{Semantic Segmentation.} The segmentation model is trained using MMSegmentation~\cite{mmseg2020}. We align the experimental setup with existing works: for MSRS, we follow the same epoch and setup as InfMAE~\cite{liu2025infmae}; for SODA and MFNet, we adopt the same setup as PAD~\cite{zhang2023pad}; and for ADE20K, we use the same setup as MCMAE~\cite{gao2022convmae}. All experiments are conducted on 4 NVIDIA V100 GPUs. Detailed configurations are provided in \cref{tab:seg}. Additionally, we modify the $\mathcal{L}_\text{PCCL}$ for the SODA and MFNet datasets to achieve faster convergence.
\begin{equation}  
\mathcal{L}^{\text{Modality}} = \text{BCEloss}\left(\sigma(\boldsymbol{M}^{\text{Modality},A}), \boldsymbol{M}^A\right).
\label{eq:pccl1}
\end{equation} 
\begin{equation}  
\mathcal{L}^{\text{Modality}} = -\log(\boldsymbol{M}^A\odot\text{softmax}(\boldsymbol{M}^{\text{Modality},A})).
\label{eq:pccl2}
\end{equation} 

\begin{table}[ht]
    \centering
    \caption{\textbf{Hyper-parameters of pre-training.}}
    \label{tab:pretraining}
    \begin{tabular}{lc}
        \toprule
        Hyperparameters & Value \\ 
        \midrule
        Input resolution & $224\times 224$ \\ 
        Optimizer & AdamW \\ 
        Base learning rate & 1.5e-4 \\ 
        weight decay & 0.05 \\ 
        Batch size & 256 \\ 
        Learning rate schedule & Cosine decay \\ 
        Warmup epochs & 40 \\ 
        Training epochs(LoRA/Full) & 400/200 \\ 
        Augmentation & Random Crop, Random Flip \\ 
        \bottomrule
    \end{tabular}
\end{table}

\begin{table}[ht]
    \centering
    \caption{\textbf{LoRA Configuration.}}
    \label{tab:lora-config}
    \begin{tabular}{lc}
    \toprule
        LoRA config & Value \\ 
    \midrule
        Target\_modules & [fc1, qkv, fc2, proj, patch\_embed4] \\ 
        LoRA\_low\_rank & 8 \\ 
        LoRA\_alpha & 32 \\ 
        LoRA\_dropout & 0.1 \\ 
    \bottomrule
    \end{tabular}
\end{table}

\begin{table}[ht]
    \centering
    \caption{\textbf{Experiment setup for object detection.}}
    \label{tab:obd}
    \begin{tabular}{lc}
    \toprule
        Hyperparameters & Value \\ 
    \midrule
        Dataset & M3FD \\
        Modality & Infrared \\ 
        Training images & 3360 \\
        Testing images & 840 \\
        Head & MaskRCNN \\
        Input resolution & $1024\times1024$ \\ 
        Pre-training Epochs & 400 \\
        Fine-tuning steps & 85k \\ 
        Optimizer & AdamW \\ 
        Base learning rate & 8e-5 \\ 
        weight decay & 0.1 \\ 
        Optimizer momentum & $\beta_1,\beta_2=0.9,0.999$ \\ 
        Batch size & 4 \\
    \bottomrule
    \end{tabular}
\end{table}

\begin{table*}[htbp]
    \centering
    \caption{\textbf{Experiment setup for downstream semantic segmentation tasks.}}
    \label{tab:seg}
    \begin{tabular}{lccccc}
    \toprule
        Hyperparameters & \multicolumn{5}{c}{Value} \\
    \midrule
        Dataset & MSRS-IR & SODA & MFNet-IR & MSRS-RGB & ADE20K \\
        Modality & Infrared & Infrared & Infrared & Visible RGB & Visible RGB \\ 
        Training images & 1083 & 1168 & 1569 & 1083 & 20210 \\
        Testing images & 361 & 1000 & 392+393 & 361 & 2000 \\
        Head & \multicolumn{5}{c}{UperNet} \\
        Input resolution & \multicolumn{5}{c}{$512\times512$} \\ 
        Pre-training epochs & 400 & 100 & 100 & 400 & 400 \\ 
        Fine-tuning steps & 48k & 14.4k & 9.8k & 60k & 160k \\ 
        Optimizer & \multicolumn{5}{c}{AdamW} \\ 
        Base learning rate & 1e-4 & 5e-4 & 5e-4 & 1e-4 & 1e-4 \\ 
        weight decay & \multicolumn{5}{c}{0.05} \\ 
        Optimizer momentum & \multicolumn{5}{c}{$\beta_1,\beta_2=0.9,0.999$} \\ 
        Batch size & 16 & 8 & 8 & 16 & 16 \\ 
        Learning rate schedule & \multicolumn{5}{c}{Poly.} \\ 
        Minimal learning rate & \multicolumn{5}{c}{0} \\ 
        Warmup steps & 1500 & 1500 & 1000 & 1500 & 1500 \\ 
        $\alpha$,$\beta$ & 1,1 & 1,1 & 1,1 & 1,1 & 1,2 \\ 
        $\mathcal{L}_\text{PCCL}$ & Eq.(\ref{eq:pccl1}) & Eq.~(\ref{eq:pccl2}) & Eq.(\ref{eq:pccl2}) & Eq.~(\ref{eq:pccl1}) & Eq.(\ref{eq:pccl1})\\
    \bottomrule
    \end{tabular}
\end{table*}

\section{Visualization}
\label{sec:app3}

\subsection{Visualization of Similarity Matrices}
\begin{figure}[ht]
    \centering
    \includegraphics[width=1\linewidth]{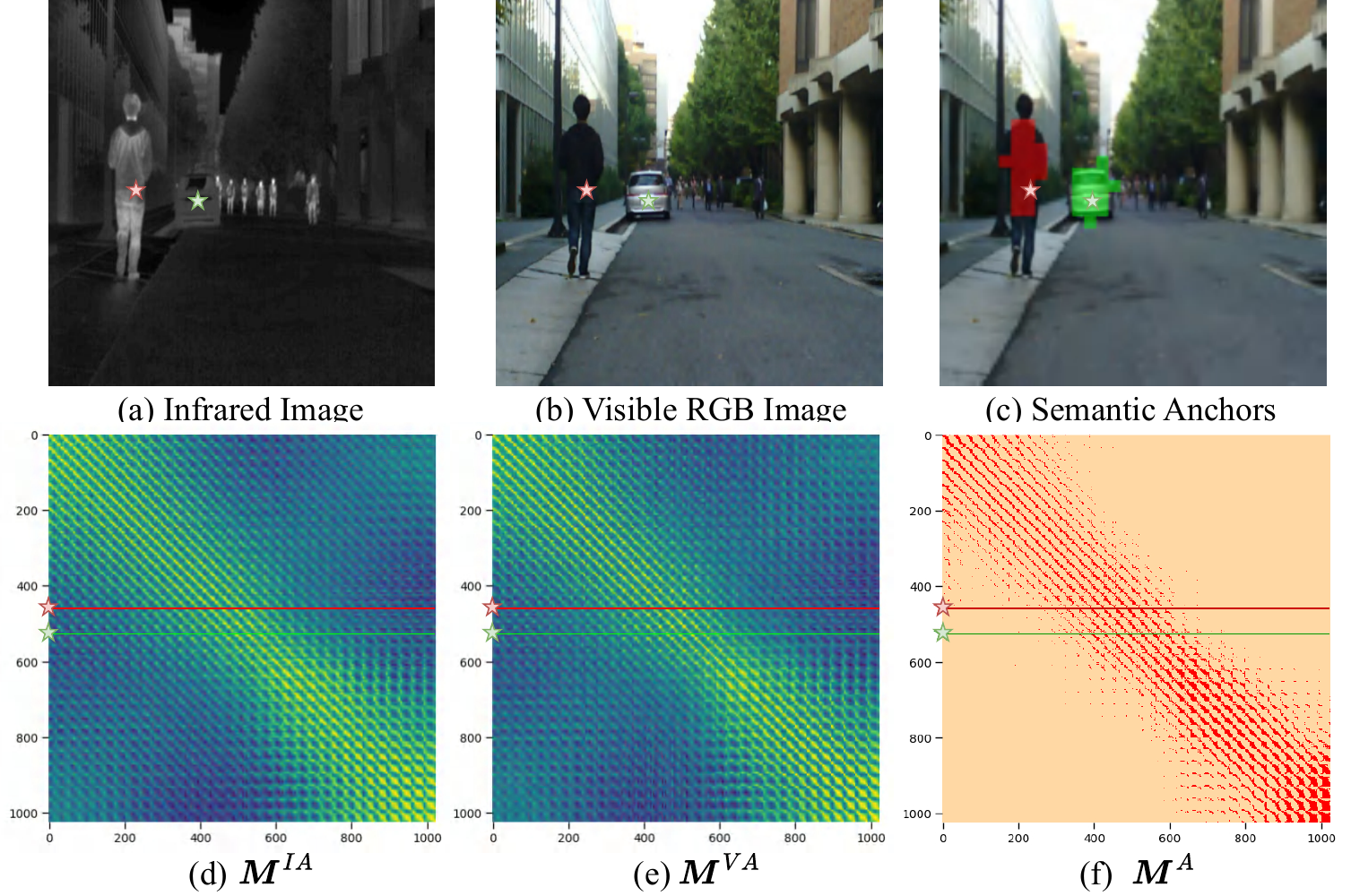}
    \caption{\textbf{Visualization of $\boldsymbol{M}^{IA}$, $\boldsymbol{M}^{VA}$, and $\boldsymbol{M}^{A}$ generated by UNIV.} Regions of interest are marked with stars. $\boldsymbol{M}^{A}$ captures the relationships between a specific patch and its surrounding patches.}
    \label{fig:simi}
\end{figure}

We present the visualization of $\boldsymbol{M}^{IA}$, $\boldsymbol{M}^{VA}$, and $\boldsymbol{M}^{A}$ in \cref{fig:simi}. The high similarity off-diagonal indicates that there are multiple related patches in a non-iconic image. \cref{fig:simi}(c) shows a clear boundary between different objects with related patches, which represents the mapping of $\boldsymbol{M}^A$'s patch attention onto the original image, highlighted with corresponding colors. The square-like pattern in \cref{fig:simi}(d-f) may be attributed to neighboring patches in the spatial domain.

\subsection{Visualization of Semantic Segmentation}
\begin{figure*}[ht]
    \centering
    \includegraphics[width=1\linewidth]{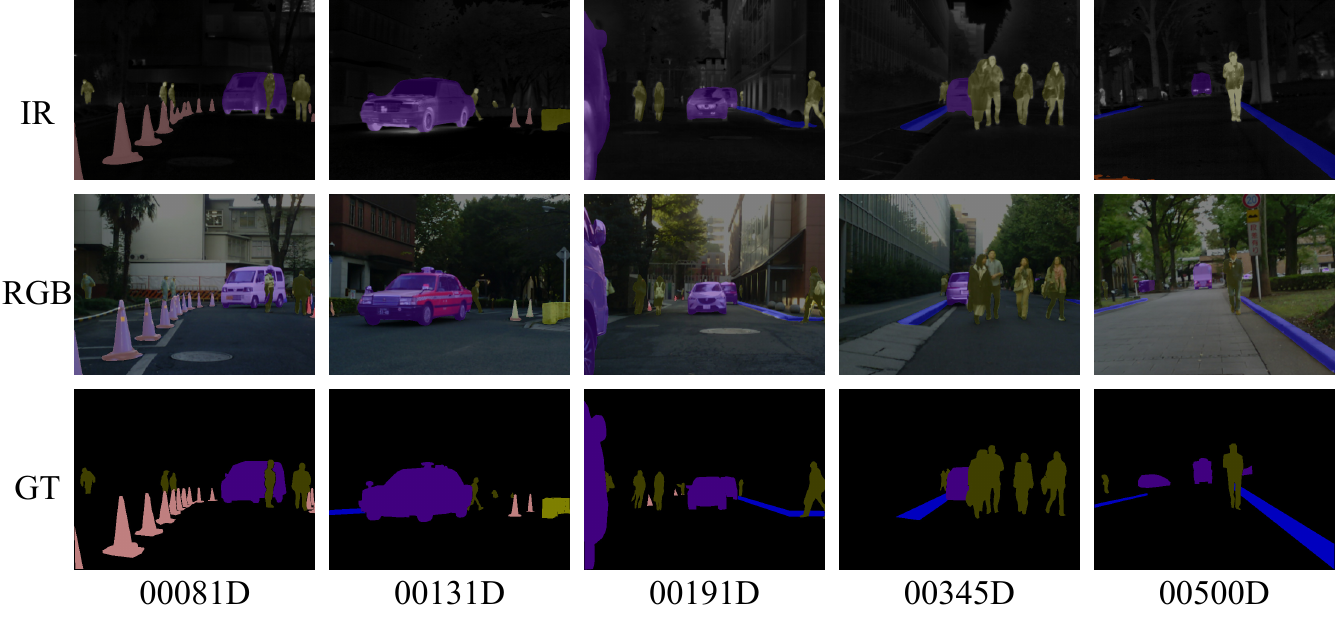}\\
    \includegraphics[width=1\linewidth]{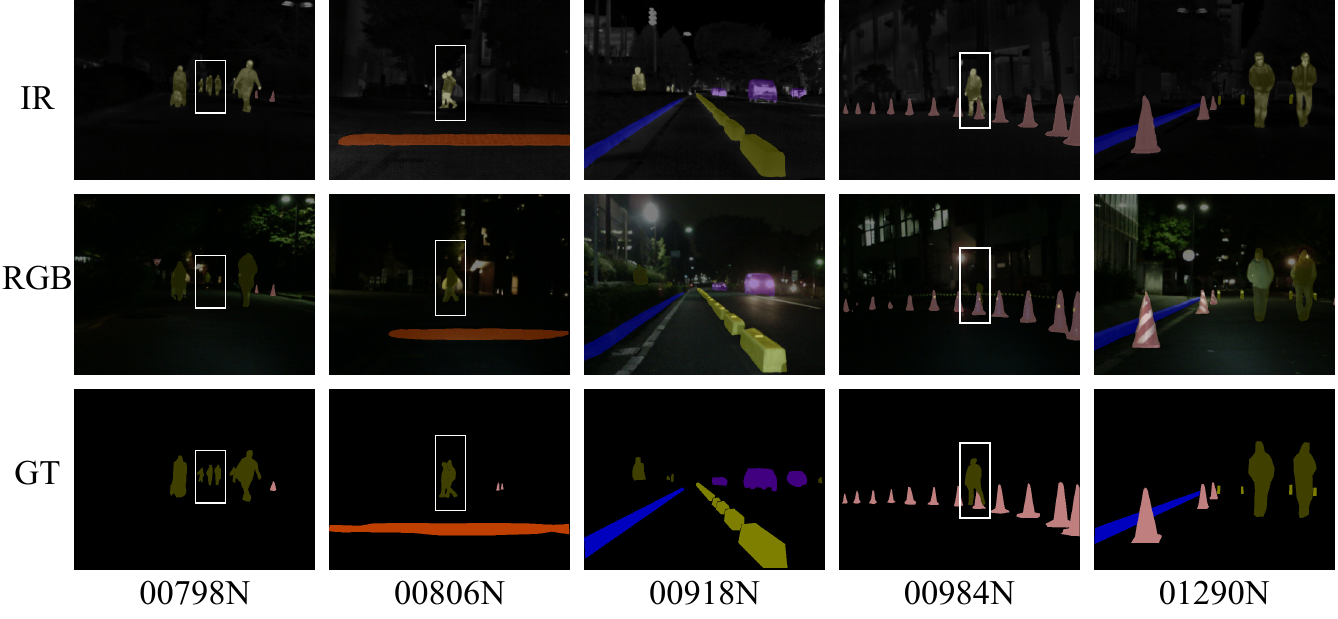}
    \caption{\textbf{Visualization of semantic segmentation on MSRS dataset.} The model is UNIV with UperNet segmentation head. Inputs are captured at both daytime and nighttime.}
    \label{fig:msrs}
\end{figure*}

The visualization results of semantic segmentation on MSRS-IR and MSRS-RGB using our UNIV framework are shown in \cref{fig:msrs}. The corresponding quantitative results are presented in Table 1 and Table 2 of the main manuscript.

Infrared images often suffer from being colorless and textureless, which hinders the detection of color-sensitive objects (e.g., color cones). Our UNIV framework leverages knowledge from the visible RGB modality to enhance the model's ability to capture features of non-heating objects, thereby narrowing the gap in detection efficiency between the two modalities. Notably, the segmentation results of nighttime images highlight the superior performance of the infrared modality. For instance, in the cases of human targets in 00798N, 00806N, and 00984N, the RGB modality fails to generate accurate semantic masks due to low-light conditions. In contrast, the infrared modality, which excels at highlighting heat-emitting objects (e.g., humans), proves more effective in detecting such targets. This phenomenon not only underscores the potential of infrared perception but also demonstrates the effectiveness of our proposed UNIV framework.

\section{Ablation Studies}
\label{sec:app4}

\subsection{LoRA Rank}
To evaluate the impact of trainable parameters on model performance, we conduct an ablation study on the rank of LoRA adaptation matrices. As shown in \cref{tab:lora-rank}, increasing the rank from 8 to 16 does not improve downstream task performance, indicating that a rank of 8 provides sufficient model complexity to capture task-relevant features. Further increasing the rank may introduce redundant parameters, potentially leading to over-fitting or reduced generalization. These results demonstrate that low-rank adaptations, such as LoRA, can effectively fine-tune models with a minimal number of parameters.

\begin{table}[htbp]
\caption{\textbf{Ablation study on LoRA rank.} $r$ represents the rank of fine-tuning matrices $\boldsymbol{A}$ and $\boldsymbol{B}$.}
\vspace{0.2cm} 
\centering
\begin{tabular}{cccc}
\toprule
    \multirow{2}{*}{\textbf{LoRA rank}} & \multirow{2}{*}{\makecell[c]{\textbf{Training}\\\textbf{Params(M)}}} & \multicolumn{2}{c}{\textbf{MSRS-IR}}\\
    \cmidrule{3-4}
    ~ &  & \textbf{mIoU} & \textbf{mAcc} \\ \midrule
    $r=8$ & 1.81 & 76.0 & 84.6 \\ 
    $r=16$ & 3.63 & 75.9 & 84.5
    \\ \bottomrule
\end{tabular}
\label{tab:lora-rank}
\end{table}

\subsection{Generalization on More Backbones}
To evaluate the generalization capability of the PCCL method across different model architectures, we conduct pre-training using MAE (ViT-L), followed by segmentation on the downstream MSRS dataset. As shown in \cref{tab:backbone}, the PCCL method effectively distills visible RGB modality knowledge across various model scales and architectures, leading to improvements in both mIoU and mAcc.
\begin{table}[htbp]
\setlength\tabcolsep{2pt}
\caption{\textbf{Ablation study on different backbones.}}
\vspace{0.2cm} 
\centering
\begin{tabular}{ccccc}
    \toprule
        \multirow{2}{*}{\textbf{Backbone}} & \multirow{2}{*}{\textbf{PCCL}} & 
        \multirow{2}{*}{\makecell[c]{\textbf{Training}\\\textbf{Params(M)}}} & \multicolumn{2}{c}{\textbf{MSRS-IR}}\\
    \cmidrule{4-5}
        \multirow{2}{*}{MAE(ViT-L)} & ~ & 304 & 75.6 & 83.2 \\
        & \CheckmarkBold & 3.16 & \textbf{75.8} (\textcolor{blue}{+0.2}) & \textbf{83.7}(\textcolor{blue}{+0.5}) \\ 
        \bottomrule
\end{tabular}
\label{tab:backbone}
\end{table}

\subsection{Hyper-parameters}
We conduct an ablation study to evaluate the impact of hyper-parameters $\gamma$ (pseudo-label selection threshold) and $\tau$ (similarity matrix sharpness). As shown in \cref{tab:hyper-parameter}, the model achieves optimal performance (mIoU: 76.0, mAcc: 84.6) with $\gamma=0.6$ and $\tau=0.04$. A larger $\gamma$ reduces pseudo-label discriminability, selecting too many patches and losing reliability, while a smaller $\gamma$ overly restricts patch relationships, hindering inter-patch correlation learning. Extreme $\tau$ values (too small or large) cause overly sharp or flat similarity distributions, impairing relevant patch extraction from $\boldsymbol{M}^{IA}$ and $\boldsymbol{M}^{VA}$. These results highlight the importance of appropriate hyper-parameter selection.

\begin{table}[htbp]
\caption{\textbf{Ablation study on hyper-parameters $\gamma$ and $\tau$.}}
\label{tab:hyper-parameter}
\vspace{0.2cm} 
\centering
\begin{tabular}{cccc}
\toprule
    \multirow{2}{*}{\textbf{Threshold}} & \multirow{2}{*}{\textbf{Temperature}} & \multicolumn{2}{c}{\textbf{MSRS-IR}} \\
    \cmidrule{3-4}
    ~ & ~ & \textbf{mIoU} & \textbf{mAcc} \\ \midrule
    $\gamma=0.3$ & $\tau=0.02$ & 75.5 & 84.0 \\
    $\gamma=0.6$ & $\tau=0.02$ & \underline{75.7} & 83.8 \\ 
    $\gamma=0.3$ & $\tau=0.04$ & 75.6 & \underline{84.1} \\ 
    $\gamma=0.6$ & $\tau=0.04$ & \textbf{76.0} & \textbf{84.6} \\ 
    \bottomrule
\end{tabular}
\end{table}

\begin{table*}[!ht]
\setlength\tabcolsep{2pt}
\caption{\textbf{Ablation study on the impact of pretraining with nighttime data.}}
\vspace{0.2cm} 
\centering
\begin{tabular}{ccccc}
    \toprule
    \multirow{2}{*}{\textbf{Backbone}} & \multirow{2}{*}{\textbf{Pretarin Dataset}} & \multirow{2}{*}{\textbf{Nighttime Data}} & \multicolumn{2}{c}{\textbf{MSRS-IR}} \\
    \cmidrule{4-5}
     &  &  & \textbf{mIoU} & \textbf{mAcc} \\
    \midrule
    \multirow{2}{*}{\textbf{UNIV}} & \multirow{2}{*}{FLIR~\cite{FLIR_Align_github}+KAIST~\cite{hwang2015multispectral}} & w/o & 75.15 & 83.44 \\
    & & w & \textbf{75.8} (\textcolor{blue}{+0.65}) & \textbf{84.29} (\textcolor{blue}{+0.85}) \\ 
    \bottomrule
\end{tabular}
\label{tab:nightCompare}
\end{table*}

\begin{table*}[!ht]
\setlength\tabcolsep{7pt}
\caption{\textbf{Model computational complexity.}}
\label{tab:real_deloyment}
\vspace{0.2cm} 
\centering
\begin{tabular}{ccccc}
    \toprule
    \textbf{Backbone} & \textbf{Segmentation Head} & \textbf{Resolution} & \textbf{TFLOPs} & \textbf{mIoU(MSRS \ IR\ /\ RGB)} \\
    \midrule
    \textbf{MCMAE}~\cite{gao2022convmae} & \multirow{3}{*}{UperNet~\cite{xiao2018unified}} & \multirow{3}{*}{$512\times512$} & 0.59 & \underline{74.9}\ /\ \textbf{79.0} \\
    \textbf{InfMAE}~\cite{liu2025infmae} & &  & 0.59 & 74.3\ /\ 75.2 \\
    \textbf{UNIV} & &  & 0.59 & \textbf{76.0}\ /\ \textbf{79.0}\\
    \bottomrule
\end{tabular}
\end{table*}

\subsection{Robustness of RGB Pretraining under Adverse Conditions}

In our study, we investigate whether incorporating nighttime scene images benefits or hinders the pretraining process. To examine this, we conduct an ablation study on the effects of low-light nighttime data inclusion. As shown in \cref{tab:nightCompare}, we evaluate the semantic segmentation performance on MSRS-IR dataset comparing backbone models trained with and without nighttime data. Our experiments reveal a counterintuitive finding: \textbf{augmenting the efficient daytime RGB dataset with nighttime data still improves the pretrained model's performance}.

We attribute this improvement to two key factors. First, nighttime scenes are not entirely devoid of illumination; most real-world collection scenarios retain sufficient light to activate the RGB knowledge network (functioning similarly to cone cells), thereby preserving meaningful visual semantics. Second, even in extreme cases (e.g., glare or heavy fog) where RGB modalities degrade, the RGB teacher model—despite occasional failures—still provides structured semantic guidance through its high-level feature representations. This enables the infrared (IR) branch of the pretrained model to develop robust high-level abstraction capabilities, generalizing effectively to IR data in challenging conditions.

Based on these insights, we integrate nighttime datasets into our pretraining strategy to enhance robustness.

\subsection{Computational Complexity}
As shown in \cref{tab:real_deloyment}, we evaluate the computational complexity of our pre-trained model. Our model achieves a balance between performance computational complexity compared to other foundation models.

{
    \small
    \bibliographystyle{ieeenat_fullname}
    \bibliography{main}
}


\end{document}